\documentclass[11pt]{article}
\usepackage[preprint]{acl}
\usepackage[T1]{fontenc}
\usepackage{newtxtext,newtxmath}
\usepackage{microtype}
\usepackage{graphicx}
\usepackage{booktabs}
\usepackage{enumitem}
\usepackage{pgfplots}
\pgfplotsset{compat=1.18}
\usepackage{tikz}
\definecolor{navy}{HTML}{004488}
\definecolor{redd}{HTML}{BB5566}
\title{A Durability and Cross-Language Transfer Benchmark for a Validated Teaching-Feedback Classification Protocol}
\author{Esteban U. Vega Barajas \\ Universidad de Guadalajara}
\begin{document}
\maketitle
\begin{abstract}
Institutions collect far more open-ended teaching-evaluation feedback than they read. A prior study introduced a validated protocol for classifying such comments by thematic category and sentiment, built from a documented annotation guide, an intra-annotator reliability measurement, stratified cross-validation, and a held-out evaluation on a Spanish institutional corpus with a frozen-encoder design. Two questions limit its reuse: whether a protocol fixed to 2019-era frozen embeddings stays competitive as representation methods advance, and whether it transfers to a second language. We re-run it on the original Spanish data across three representation generations, sparse lexical features, frozen transformer embeddings, and prompted large language models, and transfer its sentiment task to English with a balanced 45{,}000-comment corpus checked against an aspect-labeled education dataset. Treating paired comparisons as descriptive, we find the protocol durable: a 2026 frontier model posts the highest thematic F1 on the hardest Spanish task, yet shows no sentiment advantage over a cheap model and no descriptive separation from it on English, so model choice is a deployment decision, not a property of the method.

\begin{center}\large\textbf{Resumen}\end{center}
Las instituciones recogen mucho m\'as texto de evaluaci\'on docente del que llegan a leer. Un estudio previo present\'o un protocolo validado para clasificar esos comentarios por categor\'ia tem\'atica y sentimiento, construido a partir de una gu\'ia de anotaci\'on documentada, una medici\'on de fiabilidad intra-anotador, validaci\'on cruzada estratificada y una evaluaci\'on sobre un conjunto reservado en un corpus institucional en espa\~nol. Dos preguntas limitan su reutilizaci\'on: si un protocolo fijado a representaciones congeladas de la generaci\'on de 2019 sigue siendo competitivo a medida que avanzan los m\'etodos de representaci\'on, y si se transfiere a un segundo idioma. Lo reejecutamos sobre los datos originales en espa\~nol a trav\'es de tres generaciones de representaci\'on, rasgos l\'exicos dispersos, representaciones congeladas de transformadores y modelos de lenguaje grandes, y transferimos su tarea de sentimiento al ingl\'es con un corpus balanceado de 45{,}000 comentarios contrastado con un conjunto educativo etiquetado por aspectos. Tratando las comparaciones pareadas como descriptivas, encontramos que el protocolo es duradero: un modelo de frontera de 2026 obtiene el F1 tem\'atico m\'as alto en la tarea m\'as dif\'icil en espa\~nol, pero no muestra ventaja en sentimiento sobre un modelo econ\'omico ni separaci\'on descriptiva frente a \'el en ingl\'es, de modo que la elecci\'on del modelo es una decisi\'on de despliegue y no una propiedad del m\'etodo.
\end{abstract}

\noindent\textbf{Palabras clave:} evaluaci\'on docente; clasificaci\'on de texto; an\'alisis de sentimiento; transferencia transling\"ue; modelos de lenguaje grandes; PLN educativo; benchmark reproducible
\vspace{0.35em}

\section{Introduction}
Open-ended comments on teaching-evaluation surveys are among the most informative and least-used data that educational institutions hold. A single mid-sized faculty generates tens of thousands of free-text comments per academic cycle; the volume routinely exceeds what staff can read, so institutions fall back on the accompanying numeric ratings and the qualitative signal is archived unread. Natural language processing (NLP) can recover that signal, classifying each comment by what it is about and by its affective stance. For an institution to \textit{act} on the output, the analytical procedure has to be documented, reliability-checked, and validated. An accuracy number on a leaderboard does not by itself make the procedure something an institution can rely on.

Prior work \citep{vega2026} established such a procedure on a Spanish-language institutional corpus: a written annotation guide over four thematic categories and three sentiment classes; an explicit reliability measurement; stratified five-fold cross-validation for model selection; and a held-out test set untouched until the end. The reported classifiers reached weighted F1 of 0.80 for thematic classification (macro 0.75) and 0.85 for sentiment, using a Spanish BERT model (BETO; \citealp{canete2020}) as a \textit{frozen} feature extractor feeding classical classifiers. The prior work chose that design under three constraints: a small labeled set, limited compute, and the need for an auditable pipeline. Peak accuracy was not the goal.

That result invites two objections, and this paper is organized around answering them.

\paragraph{Objection 1: the method looks dated.} The pipeline freezes a 2019-era encoder and uses linear classifiers. Representation methods have since moved from sparse lexical features through contextual transformer embeddings to instruction-tuned large language models (LLMs). Is a frozen-2019 design still competitive, or has it been superseded?

\paragraph{Objection 2: it was only shown in one language.} The protocol was validated on Spanish only. Does it transfer to English educational text, and what exactly transfers?

We address both by holding the \textit{protocol} fixed and varying the \textit{model} and the \textit{language}. Concretely, we re-run the identical annotation guide, validation procedure, and evaluation on the original Spanish data across three representation generations, and we transfer the sentiment task to English using a balanced corpus sampled from a large public collection of teaching-evaluation reviews, validated against an independently annotated aspect-based education dataset.

Our framing follows the prior work: the contribution is a \textit{protocol}, not a model, and the test of a protocol is whether it holds up when the model changes and when the language changes. If a modern LLM wins on raw accuracy, that is an expected finding, and it reinforces the claim that what an institution carries forward is the documented procedure: the annotation guide, the reliability measurement, the validation design, and the interpretability analysis. None of these depends on which model is used.

\paragraph{Contributions.}
\begin{enumerate}[leftmargin=1.4em,itemsep=2pt,topsep=2pt]
\item A \textbf{durability benchmark}: the same validated protocol re-run on a single Spanish teaching-feedback corpus across three representation generations, namely sparse lexical features, frozen transformer embeddings, and prompted LLMs, with cost, latency, and auditability reported alongside F1.
\item A \textbf{cross-language transfer evaluation} of the sentiment task into English, with an explicit, narrow claim boundary and a documented, independently checked star-to-sentiment labeling rule.
\item A \textbf{reproducible, balanced English sentiment sample and its build recipe}, drawn and documented from public teaching-evaluation reviews. It holds 45{,}000 comments across three sentiment classes with fixed splits and a fixed seed. We release the build recipe, namely the script, the seed, and the split indices, so the sample is exactly reproducible, with its provenance and license stated.
\end{enumerate}

\section{Related Work}
\paragraph{NLP for student-evaluation-of-teaching (SET) text.} Automatic analysis of open-ended SET comments spans classical sentiment pipelines and, more recently, transformer classifiers. Surveys of sentiment analysis for formative assessment in higher education map the field \citep{grimalt2024}, and a broader review of NLP for educational feedback identifies sentiment annotation, topic modeling, and summarization as its dominant tasks \citep{shaik2022}. Early systems summarized free-text course feedback through unsupervised sentiment scoring and topic modeling \citep{cunningham2019}. Our protocol instead commits to supervised, reliability-checked classification, so that every label is auditable against a written guide. The structural reference closest in kind to our task is the student-feedback dataset-and-baseline work of NLPC-UOM \citep{herath2022}, which pairs an annotated teaching-feedback corpus with classical and neural baselines. Our Spanish protocol mirrors its progression from dataset to annotation, agreement, baselines, and results.

\paragraph{The nearest system: SETSum, and how we differ.} The most directly comparable effort is SETSum \citep{hu2022}, a system-demonstration pipeline that extracts aspects, scores sentiment, and summarizes institutional SETs into a reviewer-facing tool over a large English research-university corpus. We differ from it in three concrete ways. First, on population and language: SETSum operates on English research-university evaluations, whereas our protocol is validated on Spanish institutional data and is here transferred bilingually. Second, on the kind of contribution: SETSum ships a tool, while we contribute a validated methodology and test its durability. That methodology comprises a documented annotation guide, an intra-annotator reliability measurement, a controlled model comparison, and an interpretability-for-audit step. Third, on evaluation stance: we ask whether the procedure holds when the model generation changes and when the language changes, a question a tool demonstration does not address. We cite SETSum as the nearest prior work and position our benchmark against it.

\paragraph{Aspect-based and education-specific resources.} EduRABSA \citep{hua2025} provides 6{,}500 manually annotated education reviews with aspect-opinion-category-sentiment quadruplets and a review-level sentiment label, drawn from course, teacher, and university reviews. Its entity taxonomy of Course, Staff, and University gives a principled target for mapping thematic categories across datasets, and its review-level human labels serve as an independent check on our star-derived English labels. SIGHT \citep{wangR2023} contributes 15{,}784 comments over 288 MIT OpenCourseWare math-lecture transcripts with a feedback-type rubric. It is real and openly licensed, yet it sits in a public-comment register of YouTube comments rather than an institutional SET register, which bounds how we use it.

\paragraph{Cross-lingual transfer of text classification.} Whether a classifier built in one language carries to another is an empirical question, and the cross-lingual NLP literature treats it as one to be measured rather than assumed \citep{conneau2018}. Multilingual encoders pretrained jointly on many languages learn a largely shared representation space, and scaling that pretraining yields consistent gains on cross-lingual transfer tasks \citep{conneau2020}. Language-agnostic sentence embeddings push this further, letting a model built in one language apply in another with no target-language labels \citep{artetxe2019}. For sentiment specifically, multilingual transformer models transfer across languages well enough to serve as competitive off-the-shelf classifiers \citep{barbieri2022}. These results motivate a single multilingual encoder as one arm of our benchmark, and they frame Experiment~B as a re-validation of the protocol in English rather than a presumption that the Spanish result carries over.

\paragraph{LLMs as zero- and few-shot classifiers.} Large language models perform many tasks from a prompt and a handful of demonstrations, with no parameter updates \citep{brown2020}. For text-as-data work this makes them candidate substitutes for human coders: instruction-tuned models match or exceed crowd annotators on several annotation tasks \citep{gilardi2023}, and a broad survey maps both their promise and their limits as classifiers for computational social science \citep{ziems2024}. Follow-up work tempers this: model annotations can be unstable and sensitive to prompt wording \citep{reiss2023}, and using them as labels calls for validation against human annotation \citep{pangakis2023}. The picture for sentiment is more measured. A systematic check finds that LLMs handle simple sentiment well but trail smaller, domain-trained models on harder structured sentiment \citep{zhang2024}. Our durability benchmark tests that limit directly on teaching-feedback text.

\paragraph{Representation methods.} The three arms of our durability benchmark correspond to three representation generations. The first is sparse lexical features, term frequency-inverse document frequency (TF-IDF), with linear classifiers. The second is mean-pooled frozen transformer embeddings \citep{reimers2019} over BERT-family encoders \citep{devlin2019} such as BETO \citep{canete2020}; the frozen-versus-fine-tuned trade-off for small labeled survey data has been analyzed by \citet{gweon2024}. The third is instruction-tuned LLMs prompted zero- and few-shot against the same label definitions. Holding the protocol constant across these arms isolates its contribution from the model's.

\section{Protocol and Data}
\subsection{The carried-over protocol}
We re-use, without modification, the protocol from the prior Spanish study, which has four parts. A written annotation guide defines four thematic categories: \textit{methodology}; \textit{evaluation}, meaning evaluation practices; \textit{interaction}, meaning student-educator treatment; and \textit{attendance/engagement}. The guide also defines three sentiment classes: \textit{positive}, \textit{negative}, and \textit{neutral}. Reliability is measured as \textbf{test-retest intra-annotator} agreement, with a single annotator re-labeling a 100-comment sample one week apart. This yields Cohen's $\kappa = 0.82$ for thematic and 0.88 for sentiment, values that estimate the stability of one annotator over time and that are \textbf{not} inter-annotator agreement. Model selection uses stratified five-fold cross-validation. A held-out test set of 20\% is reserved for final reporting only. The same guide and procedure apply to every arm and to both languages, so the comparison tests the protocol, not a particular model.

\subsection{Spanish data}
\label{sec:esdata}
The Spanish evaluation uses the original institutional corpus: 44{,}707 valid open-ended comments from the SIIAU teaching-evaluation system at Universidad de Guadalajara, collected across three cycles, 2022B, 2023A, and 2023B, and anonymized. A labeled subset of 1{,}165 comments was annotated by a single annotator, and a 20\% held-out set of 233 comments is reserved for final reporting. In that subset the sentiment classes are imbalanced: negative is the largest at 46.5\%, positive follows at 39.7\%, and neutral is smallest at 13.7\%. All Spanish supervised results derive from this gold-labeled subset, and we cite the thesis \citep{vega2026} to declare the derivation. We are separately assembling a further institutional corpus in Spanish that spans 2014 to 2025 and is de-identified. That corpus is unlabeled and does not enter the supervised results; annotating it for temporal and recency validation is future work.

\subsection{English data}
\label{sec:endata}
The English transfer uses three openly-licensed resources.

\paragraph{Primary sentiment corpus, assembled here.} We sample a balanced 45{,}000-comment sentiment corpus from a large public collection of RateMyProfessor teaching-evaluation reviews released for research \citep{he2020,he2022}. The full release used here is licensed CC BY-NC-SA per its Baidu AI Studio release page; the smaller Mendeley sample carries the more permissive CC BY 4.0 license. From 9.54M reviews we derive a sentiment label from the per-review star rating, under the same three-class scheme as the Spanish protocol: a review is positive if its star rating is at least 4, negative if it is at most 2, and neutral otherwise. We then clean the text and balance the classes. We decode escaped byte sequences, drop comments under 15 characters and exact duplicates, and remove reviews without a star rating. To avoid the file's professor and school ordering, we draw a class-balanced sample by uniform per-class reservoir sampling over the full file at a fixed seed, giving 15{,}000 comments per class. The stratified train, development, and test split holds 31{,}500, 6{,}750, and 6{,}750 comments. Mean comment length is 231 characters. We release the build script, the seed, and the split indices so the sample is exactly reproducible; the underlying review text stays governed by the source CC BY-NC-SA license. One variant of the collection is excluded by design: a moral-foundations release circulates with a \texttt{student\_star} column truncated to the 4.0 to 5.0 range, which cannot yield a balanced sentiment label.

\paragraph{Independent gold check.} EduRABSA's review-level human sentiment labels \citep{hua2025} provide a non-heuristic test set against which we validate the star-to-sentiment rule, and we report the agreement between star-derived and human labels.

\paragraph{Register-contrast resource.} SIGHT \citep{wangR2023} is used only where a non-institutional, public-comment register is informative for the discussion of register gaps; it is not used to make institutional-SET claims.

\paragraph{Excluded by design.} We exclude the synthetic, LLM-generated ``student feedback'' datasets (e.g., Daye34, Philseok) as primary corpora: validating a methodology meant for \textit{real} feedback on machine-generated text would be a validity error. Where such data appear at all, it is only as a clearly-labeled stress test.

\subsection{Star-to-sentiment labeling and its validation}
\label{sec:starlabel}
Star-derived labels are a convenience signal, not ground truth, and the literature on course-review data is explicit that the mapping must be documented and validated. We state the rule above and measure its agreement with EduRABSA's human review-level sentiment. We report English results with the labeling-noise caveat attached. We therefore test transfer of a \textit{procedure} under a labeling rule whose error we measure against human labels. On the 3{,}000 RateMyProfessor teacher reviews within EduRABSA, the subset that shares the star-rating scale of our corpus, the star rule agrees with human review-level sentiment at Cohen's kappa 0.66, with 82\% accuracy. The neutral class is the weakest, because human-neutral reviews scatter across star levels. The English labels are therefore a reasonable but imperfect proxy. This kappa measures agreement between star-derived and human labels, and it is not the thesis's intra-annotator kappa.

\section{Experimental Setup}
\label{sec:setup}
\paragraph{Model arms: same protocol, three generations.}
\begin{enumerate}[leftmargin=1.4em,itemsep=2pt,topsep=2pt]
\item \textbf{Sparse-lexical anchor.} TF-IDF features \citep{sparckjones1972} with a linear support vector classifier \citep{cortes1995}, the continuity baseline carried from the prior study.
\item \textbf{Frozen-embedding 2019-generation arm.} Mean-pooled frozen transformer embeddings, 768-dimensional, feeding the prior study's classifiers: LinearSVC (\texttt{C = 0.1}, \texttt{dual = False}) for thematic, and SVC-RBF with \texttt{class\_weight = balanced} for sentiment. Spanish uses BETO (\texttt{dccuchile/\allowbreak{}bert-base-spanish-\allowbreak{}wwm-uncased}). BETO is specific to Spanish, so for English the modern multilingual-encoder arm below is the representative frozen-embedding generation.
\item \textbf{Modern multilingual-encoder arm.} Mean-pooled frozen embeddings from a current multilingual encoder, the \texttt{intfloat/\allowbreak{}multilingual-e5-base} model \citep{wangL2024}, feeding the same classifiers. This arm isolates the effect of a newer representation under an unchanged protocol.
\item \textbf{LLM arms.} Instruction-tuned LLMs prompted zero-shot and few-shot with the annotation guide's label definitions, evaluated on the identical held-out items, at three points: a cheap hosted model, Claude Haiku 4.5 \citep{anthropicHaiku2025}; the current frontier, Claude Opus 4.8 \citep{anthropicOpus2026}; and a small open-weights model run locally, Qwen2.5-1.5B-Instruct \citep{qwen2024}. We compare them to isolate whether frontier capability, holding the documented protocol fixed, drives accuracy, and whether an on-premise open model is viable. The first two are closed API models. The third is the on-premise and reproducible option, evaluated on Spanish on CPU and on English on a consumer GPU.
\end{enumerate}
An optional parameter-efficient fine-tuning arm is a staged extension and is out of scope here.

\paragraph{Selection and evaluation.} Stratified five-fold cross-validation selects classical-arm hyperparameters before any contact with the held-out set. We report weighted F1 as the primary metric and macro F1 to expose class imbalance, with per-class precision, recall, and F1. Because the classes are imbalanced, accuracy is not the headline metric.

\paragraph{Model comparison, descriptive only.} Paired comparisons are reported as descriptive evidence of consistency, \textbf{not as significance tests}. We use McNemar's test on held-out predictions \citep{mcnemar1947,dietterich1998}, with descriptive bootstrap confidence intervals around F1 differences reported in Appendix~\ref{app:boot}. We make \textbf{no claim of statistical significance}. In the prior study these comparisons favored the proposed models in the expected direction but did not reach the conventional threshold, for example $p = 0.512$ for sentiment versus the \texttt{pysentimiento} baseline \citep{perez2021}, and we carry that discipline forward.

\paragraph{Interpretability and cost.} LIME (Local Interpretable Model-agnostic Explanations; \citealp{ribeiro2016}) is applied to a sample of predictions as an auditability sanity check (Appendix~\ref{app:lime}). For each arm we also report inference cost, latency, and auditability; for institutional adoption these enter the model-selection decision.

\section{Experiment A: Spanish durability across model generations}
\label{sec:expA}
This experiment asks whether the protocol's performance holds as the representation is modernized. All arms use the identical labeled subset, splits, guide, and held-out set.

\begin{table*}[t]\centering\footnotesize\setlength{\tabcolsep}{4.5pt}
\begin{tabular}{l rrr l l}
\toprule
Arm & Thematic wF1 & Thematic macro F1 & Sentiment wF1 & Cost / latency & Auditability \\
\midrule
TF-IDF + LinearSVC & 0.602 & 0.459 & 0.814 & low / low & high (sparse, LIME) \\
Frozen BETO-2019 & 0.744 & 0.681 & 0.838 & low / low & high (LIME) \\
e5 encoder & 0.740 & 0.678 & 0.859 & medium & medium \\
\midrule
Haiku 4.5 zero-shot & 0.854 & 0.832 & \textbf{0.923} & high / high & low (prompt-bound) \\
Haiku 4.5 few-shot & 0.841 & 0.817 & 0.922 & high / high & low \\
Opus 4.8 zero-shot & 0.874 & 0.854 & 0.884 & highest / high & low (prompt-bound) \\
Opus 4.8 few-shot & \textbf{0.886} & \textbf{0.864} & 0.910 & highest / high & low \\
Qwen2.5-1.5B zero-shot & 0.307 & 0.354 & 0.761 & zero API / high & medium (open) \\
Qwen2.5-1.5B few-shot & 0.626 & 0.567 & 0.819 & zero API / high & medium (open) \\
\bottomrule
\end{tabular}
\caption{\textbf{Spanish durability results by arm} on the 233-comment held-out set (single stratified split, seed 20260616). Best value per metric column in bold; the rule separates the non-LLM arms from the prompted LLMs. Figures are descriptive; no significance is claimed.}
\label{tab:durability}
\end{table*}

\paragraph{Anchored result against the prior study.} The frozen-embedding arm with classical classifiers reached the prior study's reported weighted F1 0.80 thematic (macro 0.75) and 0.85 sentiment on its original split. On that split it reached higher weighted F1 than an internal TF-IDF and LinearSVC baseline on thematic and than the published \texttt{pysentimiento} baseline on sentiment. This is a descriptive F1 comparison, and the paired tests did not reach significance (Section~\ref{sec:setup}). The \textit{neutral} sentiment class and the sparsely-populated \textit{attendance/engagement} thematic class show the lowest per-class F1, as expected for a residual class and a rare class respectively (Table~\ref{tab:perclass}).

\begin{figure}[t]\centering
\begin{tikzpicture}
\begin{axis}[
  width=7.0cm, height=5.7cm,
  axis lines=left, axis line style={-, gray!55},
  ymin=0.25, ymax=1, ytick={0.25,0.5,0.75,1.0}, yticklabels={0.25,0.50,0.75,1.00},
  xmin=-0.5, xmax=8.5,
  xtick={0,1,2,3,4,5,6,7,8},
  xticklabels={TF-IDF,BETO,e5,Haiku-0,Haiku-f,Opus-0,Opus-f,Qwen-0,Qwen-f},
  x tick label style={rotate=40, anchor=east, font=\scriptsize},
  yticklabel style={font=\scriptsize},
  ylabel={Weighted F1}, ylabel style={font=\footnotesize},
  ymajorgrids, grid style={gray!18, line width=0.4pt},
  tick align=outside, clip=false,
  legend style={at={(0.5,1.12)}, anchor=south, legend columns=2, font=\scriptsize, draw=gray!40, column sep=1ex},
  legend cell align=left,
]
\draw[gray!45, dashed, line width=0.5pt] (axis cs:0.5,0.25) -- (axis cs:0.5,1);
\draw[gray!45, dashed, line width=0.5pt] (axis cs:2.5,0.25) -- (axis cs:2.5,1);
\addplot[navy, line width=1.3pt, mark=*, mark size=2.2pt, mark options={fill=navy, draw=white, line width=0.5pt},
  error bars/.cd, y dir=both, y explicit, error bar style={navy!45, line width=0.5pt}]
  table[x=x,y=y,y error plus=ep,y error minus=em] {
x y ep em
0 0.6022 0.0715 0.0765
1 0.7444 0.0542 0.0576
2 0.7399 0.0553 0.0590
3 0.8539 0.0436 0.0445
4 0.8409 0.0439 0.0455
5 0.8735 0.0412 0.0427
6 0.8865 0.0383 0.0424
7 0.3074 0.0695 0.0736
8 0.6260 0.0594 0.0579
};
\addlegendentry{Thematic}
\addplot[redd, line width=1.3pt, dashed, mark=square*, mark size=2.2pt, mark options={solid, fill=redd, draw=white, line width=0.5pt},
  error bars/.cd, y dir=both, y explicit, error bar style={redd!45, line width=0.5pt}]
  table[x=x,y=y,y error plus=ep,y error minus=em] {
x y ep em
0 0.8140 0.0506 0.0557
1 0.8381 0.0455 0.0517
2 0.8592 0.0435 0.0491
3 0.9234 0.0335 0.0367
4 0.9221 0.0315 0.0312
5 0.8837 0.0390 0.0461
6 0.9102 0.0323 0.0363
7 0.7609 0.0633 0.0669
8 0.8189 0.0544 0.0569
};
\addlegendentry{Sentiment}
\end{axis}
\end{tikzpicture}
\caption{\textbf{Spanish durability by representation generation.} Weighted F1 on the 233-comment held-out set (single stratified split, seed 20260616), arms ordered by generation: sparse lexical, frozen embeddings, and large language models, separated by dashed rules. Thematic = navy solid circles, sentiment = red dashed squares; caps are descriptive 95\% bootstrap confidence intervals (no significance is claimed). Thematic F1 rises sharply at the LLMs while sentiment stays high and flat, and the open-weights Qwen2.5-1.5B dips hardest.}
\label{fig:durability}
\end{figure}
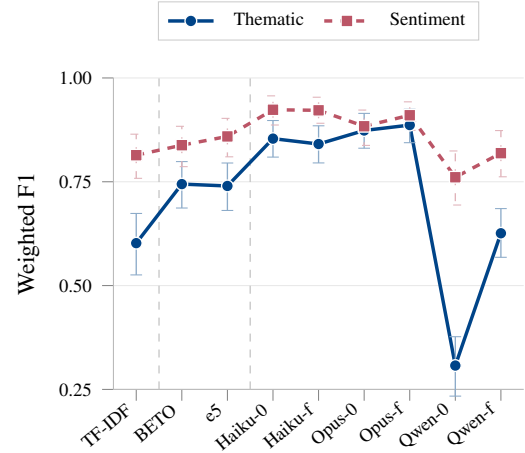

\paragraph{The unified-split re-run.} All arms share a single stratified split with seed 20260616 and a 233-comment held-out set (Table~\ref{tab:durability}, Figure~\ref{fig:durability}). These figures are distinct from the canonical thesis result cited above (thematic wF1 0.80, macro 0.75; sentiment wF1 0.85) and are reported separately because the thesis's exact split was not preserved. The frozen-BETO re-run lands slightly below the original-split figure, as expected. The durability claim rests on the cross-arm ordering under one fixed protocol, not on matching the original number.

\begin{table}[t]\centering\small
\begin{tabular}{l rrr r}
\toprule
Class & P & R & F1 & $n$ \\
\midrule
\multicolumn{5}{l}{\textit{Sentiment}} \\
Negative & 0.83 & 0.93 & 0.87 & 108 \\
Neutral & 0.46 & 0.34 & 0.39 & 32 \\
Positive & 0.98 & 0.92 & 0.95 & 93 \\
\midrule
\multicolumn{5}{l}{\textit{Thematic}} \\
Methodology & 0.83 & 0.81 & 0.82 & 136 \\
Evaluation & 0.59 & 0.65 & 0.62 & 34 \\
Interaction & 0.69 & 0.67 & 0.68 & 36 \\
Attendance/engagement & 0.59 & 0.63 & 0.61 & 27 \\
\bottomrule
\end{tabular}
\caption{\textbf{Per-class precision, recall, and F1} for the frozen BETO-2019 arm on the Spanish held-out set ($n{=}233$). The neutral sentiment class and the rare attendance/engagement thematic class are the weakest, as expected for a residual and a sparsely-populated class.}
\label{tab:perclass}
\end{table}

\paragraph{Reading the Spanish results.} Under the fixed protocol the arms form a consistent ordering. TF-IDF is weakest. Frozen BETO-2019 and the modern e5-base encoder are numerically close, with no descriptive separation: thematic McNemar $b{=}29, c{=}28$ and sentiment $b{=}12, c{=}17$, and no significance test is run. The two LLMs are strongest. The largest single jump is in thematic macro F1: it rises from 0.46 for TF-IDF to about 0.68 for BETO and the encoder, then to roughly 0.82 to 0.86 for the two hosted LLMs. The LLMs handle the rare and residual classes that the frozen-2019 design struggles with; on thematic, BETO versus Opus few-shot gives McNemar $b{=}21, c{=}54$, a large gap in the LLM's favor. The easier task behaves differently. The 2026 frontier model shows no sentiment advantage over the cheap one on this held-out set: Haiku zero-shot reaches 0.923 against Opus zero-shot at 0.884, and the head-to-head McNemar is $b{=}13, c{=}4$ in Haiku's favor. Across the full benchmark, Opus cost about 7.7x as much as Haiku in total API spend (USD 8.25 versus 1.08). We report this descriptively and claim no significance. We do not establish whether the binding constraint is the label scheme or model capability; sentiment weighted F1 already sits near 0.92 on this 233-item set, so a task or label ceiling may explain the missing frontier gain more than capability does. That ceiling sits on the easy positive and negative mass, where the neutral class stays weakest. We report all McNemar counts as descriptive consistency only. The TF-IDF versus BETO sentiment comparison is itself near-even at $b{=}17, c{=}22$, echoing the original study's non-significant result.

\paragraph{Ablation.} Holding the classifier and guide fixed and varying only the representation isolates how much of the performance is attributable to the representation generation rather than the protocol; Table~\ref{tab:durability} reports this comparison across the arms. We re-use the prior study's selection grid unchanged so the ablation is comparable across arms; that grid spans representation, balancing, and classifier family, and includes \texttt{class\_weight} and SMOTE/ADASYN variants.

\begin{figure}[tbp]\centering
\begin{tikzpicture}
\begin{axis}[
  width=7.3cm, height=5.8cm,
  axis lines=left, axis line style={-, gray!55},
  xmode=log, xmin=0.03, xmax=16, ymin=0.70, ymax=0.96,
  xlabel={API cost (USD, log scale; local arms $\approx 0$)}, xlabel style={font=\small},
  ylabel={Spanish sentiment weighted F1}, ylabel style={font=\small},
  xtick={0.05,0.1,1,10}, xticklabels={local,0.1,1,10},
  ytick={0.70,0.75,0.80,0.85,0.90,0.95},
  tick label style={font=\footnotesize},
  ymajorgrids, grid style={gray!18}, clip=false,
  legend style={at={(0.97,0.04)}, anchor=south east, font=\footnotesize, draw=gray!40, fill=white, row sep=1pt},
  legend cell align=left,
]
\addplot[gray!55, dashed, line width=0.8pt, forget plot] coordinates {(0.075,0.8592) (1.08,0.9234)};
\addplot[only marks, navy, mark=*, mark size=2.6pt, mark options={draw=white}] coordinates {(0.045,0.8140)};
\addlegendentry{TF-IDF (local)}
\addplot[only marks, navy, mark=diamond*, mark size=3.4pt, mark options={draw=white}] coordinates {(0.075,0.8592)};
\addlegendentry{e5 encoder (local)}
\addplot[only marks, gray!70, mark=triangle*, mark size=3.2pt] coordinates {(0.055,0.8189)};
\addlegendentry{Qwen few (local)}
\addplot[only marks, redd, mark=square*, mark size=2.8pt, mark options={draw=white}] coordinates {(1.08,0.9234)};
\addlegendentry{Haiku zero}
\addplot[only marks, redd, mark=diamond*, mark size=3.6pt, mark options={draw=white}] coordinates {(8.25,0.9102)};
\addlegendentry{Opus few}
\end{axis}
\end{tikzpicture}
\caption{\textbf{Cost versus accuracy on Spanish sentiment.} Each arm's weighted F1 against API cost on a log scale, where the cost coordinate for each LLM marker is that model's total API spend across the whole benchmark and local arms run at effectively zero API cost. The dashed line is the cost-accuracy frontier. Opus 4.8 sits below Haiku 4.5 despite about 7.7x the total API spend, while the frozen e5 encoder and the open-weights Qwen2.5-1.5B arm are near-free and competitive.}
\label{fig:cost}
\end{figure}
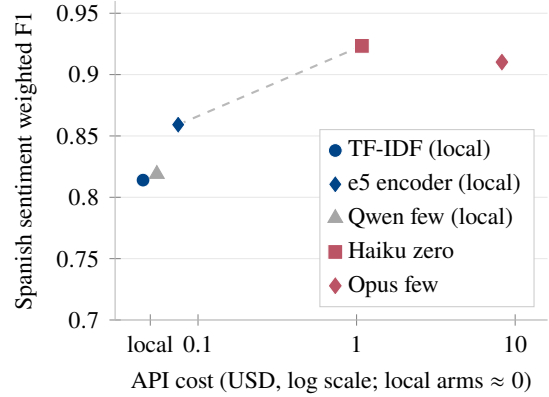

\paragraph{Durability conclusion for Spanish.} The LLMs win on raw accuracy, the expected finding, but they do not obsolete the protocol. They slot in as additional arms under the same documented annotation guide, splits, and held-out evaluation, and they win at a cost: higher inference price and latency, and lower auditability, since their decisions are prompt-bound, unlike the inspectable sparse weights and the LIME-explained classical arms. On sentiment, the 2026 frontier model buys no accuracy gain over the cheap one but costs far more, the parity-at-a-cost pattern reported above and visualized in Figure~\ref{fig:cost}. The frozen-2019 design and the modern encoder stay competitive and remain the cheapest and most auditable arms. At the opposite end, the small open-weights model Qwen2.5-1.5B is the weakest arm, especially zero-shot on the four-class thematic task, where wF1 is 0.31 and rises to 0.63 with few-shot. It runs at zero API cost and on-premise, with lower accuracy but full reproducibility and privacy. The protocol holds across all of these arms; no single arm is load-bearing.

\section{Experiment B: English transfer}
\label{sec:expB}
This experiment asks what transfers to English when the protocol is applied unchanged.

\paragraph{Sentiment transfer, primary.} We train and evaluate the same arms on the assembled 45{,}000-comment balanced English corpus (Section~\ref{sec:endata}), reporting weighted and macro F1 on the held-out test split (Table~\ref{tab:english}), plus agreement between star-derived and EduRABSA human labels. Because the classes are balanced, weighted F1 and macro F1 are essentially equal. The star-versus-human label validation against EduRABSA is reported in Section~\ref{sec:starlabel}, at Cohen's kappa 0.66 on RMP teacher reviews. No frozen-2019 English-encoder arm is included, because BETO is Spanish-specific and the modern multilingual encoder is the representative embedding generation for the English transfer.

\begin{table}[t]\centering\small
\begin{tabular}{l rr}
\toprule
Arm & wF1 & macro F1 \\
\midrule
TF-IDF with LinearSVC & 0.675 & 0.675 \\
e5 encoder & 0.729 & 0.729 \\
\midrule
Haiku 4.5 zero-shot ($n{=}1500$) & 0.671 & 0.667 \\
Haiku 4.5 few-shot ($n{=}1500$) & 0.733 & 0.730 \\
Opus 4.8 zero-shot ($n{=}1500$) & 0.672 & 0.668 \\
Opus 4.8 few-shot ($n{=}1500$) & 0.723 & 0.720 \\
Qwen2.5-1.5B zero-shot ($n{=}1500$) & 0.558 & 0.552 \\
Qwen2.5-1.5B few-shot ($n{=}1500$) & 0.620 & 0.615 \\
\bottomrule
\end{tabular}
\caption{\textbf{English sentiment transfer} on the balanced three-class test split. The classical and encoder arms use the full 6{,}750-comment split; the LLM arms use a 1{,}500-comment cost-capped subset, so values are not directly comparable across that rule and are not bolded.}
\label{tab:english}
\end{table}

\paragraph{English results.} Sentiment transfers, but the ranking differs from Spanish. On the full test split the modern encoder is the strongest arm, at wF1 0.73. Both LLMs match it only with few-shot prompting: zero-shot sits near 0.67 for Haiku and Opus alike, and few-shot reaches 0.72 to 0.73. TF-IDF trails at 0.67. The frontier model brings no English advantage over the cheap one, with Opus and Haiku within a point at every setting, and neither closed LLM leads English by the descriptive margin both show on Spanish. The open-weights model Qwen2.5-1.5B is again the weakest arm, at zero-shot 0.56 and few-shot 0.62. Two factors bound the ceiling. The English labels are star-derived and noisy, and a classifier cannot exceed the label agreement, which is Cohen's kappa 0.66 against EduRABSA human labels (Section~\ref{sec:starlabel}); the Spanish results, by contrast, use gold labels. The protocol transfers to English sentiment at a useful but lower accuracy. A frozen modern encoder is the strongest and cheapest English arm on the full test split, though its lead over the LLMs is only indicative, since the LLM figures use a smaller 1{,}500-comment subset. Few-shot examples are what lift the LLMs to encoder parity.

\paragraph{Thematic transfer.} The four Spanish thematic categories do not map one-to-one onto an English course-review taxonomy. We align them to EduRABSA's entity categories of Course, Staff, and University and report the alignment (Table~\ref{tab:align}); the transfer is a documented category mapping, not yet a solved classification.

\begin{table}[t]\centering\footnotesize\setlength{\tabcolsep}{4pt}
\begin{tabular}{@{}p{1.8cm} p{1.55cm} p{3.35cm}@{}}
\toprule
Spanish category & EduRABSA entity & Correspondence \\
\midrule
Methodology & Course, Staff & Teaching methods and course delivery map to Course attributes such as content and structure, and to Staff teaching behavior. \\
Evaluation & Course, Staff & Assessment, exams, and grading map to Course assessment attributes and to Staff grading fairness. \\
Interaction & Staff & Treatment, respect, and communication map to Staff attributes. \\
Attendance\slash engagement & Staff, Course & Attendance, participation, and engagement map to Staff engagement and to Course workload. \\
\bottomrule
\end{tabular}
\caption{\textbf{Spanish-to-EduRABSA thematic alignment.} A documented entity-level mapping, not a validated cross-lingual classifier. EduRABSA's University entity, which covers institution-level facilities and administration, is not targeted by our teaching-focused categories.}
\label{tab:align}
\end{table}

\paragraph{Claim boundary.} The English data are higher-education teaching-evaluation reviews, not K-12 family feedback. There is no public K-12 free-text corpus to test that register, and we do not claim to cover it. The transfer we evaluate is sentiment, under a star-derived labeling rule whose noise we measure against human labels (Section~\ref{sec:starlabel}). Thematic transfer is scoped to a category mapping. The English evaluation is not an equity audit and makes no demographic-subgroup claim; equity is out of scope for this paper by design.

\section{Discussion}
We slotted every newer model into the same annotation guide, splits, and held-out evaluation, and none forced the auditability or cost story to collapse. Within that fixed frame the representation and the language change around it. Where accuracy rises with newer representations, the procedure carries over unchanged, so the value sits in the procedure and not the model; where the cheap frozen design stays competitive, an auditable, low-cost pipeline already suffices. One result here is not the expected one: within the fixed protocol the 2026 frontier model, Opus 4.8, gives no sentiment gain over the cheaper Haiku 4.5 while costing far more, and the near-free frozen encoder stays on the cost-accuracy frontier. We report this descriptively and claim no significance.

\section{Conclusion}
We have built a benchmark that retests a validated teaching-feedback protocol on two fronts: whether its model is now dated, and whether it transfers beyond Spanish. We answer them by re-running the identical protocol across three representation generations and transferring it to English under an explicit claim boundary. We release a reproducible, balanced English sentiment corpus as a community artifact. Across the representation generations the protocol holds: a modern multilingual encoder matches the frozen BETO-2019 design, and the LLMs lead on Spanish, most of all on the rare thematic classes. The 2026 frontier model shows no descriptive sentiment advantage over a cheap LLM on either language's evaluated splits. In English, a frozen encoder is the strongest and cheapest arm among those scored on the full test split, with the LLM arms only indicatively comparable because they used a smaller cost-capped subset. What carries across is the procedure; the model running underneath it is a deployment choice, weighed on cost and on what an institution can audit.

\section*{Limitations}
\begin{itemize}[leftmargin=1.2em,itemsep=2pt,topsep=2pt]
\item \textbf{Single annotator.} All Spanish gold labels come from one annotator, so the $\kappa$ values are test-retest intra-annotator stability, not inter-annotator agreement. A second annotator is the most important next step.
\item \textbf{Single split.} The Spanish durability ordering is read from one stratified split with seed 20260616 and a 233-comment held-out set. We report descriptive bootstrap intervals but do not vary the split, so robustness across resampled or repeated splits is future work.
\item \textbf{Spanish and English results are not directly comparable.} The two settings differ in label source, in domain, and in training-set size: Spanish uses gold human labels, while English uses star-derived labels. The dominant confound is label quality rather than size. The LLM arms are prompted and do not train, so the larger English set gives them no advantage; the gold-versus-star-derived label gap is what separates the two settings. For the trained arms, TF-IDF and the encoder, the English training set is much larger than the Spanish one, so the size asymmetry runs in opposite directions for trained and prompted arms. A down-sampled English training set matched to the Spanish size would neutralize the size objection directly, and we leave that robustness run to future work.
\item \textbf{Star-derived English labels.} English sentiment labels are heuristic. We quantify their agreement with human labels but do not eliminate the noise.
\item \textbf{Corrupted structural metadata in the English source.} School, department, and state fields in the public RMP collection are unreliable parsing artifacts, so we make no per-institution or per-region English claims.
\item \textbf{Register and level gaps.} English data are higher-education reviews, SIGHT is public YouTube comments on lecture videos rather than institutional SET, and neither is K-12 family feedback.
\item \textbf{Topic mapping is imperfect.} Cross-language thematic transfer is a documented alignment, not a validated classifier.
\item \textbf{Provenance of the English corpus.} The RMP collection is scraped, self-selected, commercial-platform data under CC BY-NC-SA. It is appropriate for a research benchmark with provenance stated rather than for a public-institution corpus. These are public, instructor-directed consumer reviews and can name the professor; we release no raw comment text and reproduce none verbatim, so no English review text appears in the released or displayed artifacts.
\end{itemize}

\section*{Ethical Considerations}
The intended use of this work is institutional teaching improvement and the triage of large volumes of otherwise unread open-ended feedback. It is not designed for the high-stakes evaluation of individual instructors. The classifiers and prompted models studied here produce a noisy signal, and as Section~\ref{sec:starlabel} shows even the validation labels are imperfect proxies. Their output should support aggregate review and triage. It must not serve as ground truth for personnel decisions. The Spanish institutional corpus was accessed under Universidad de Guadalajara's research-use framework and is used here in anonymized form; the paper reports no raw comment text. The English data are public RateMyProfessor reviews under a CC BY-NC-SA license, used only for a research benchmark with provenance stated. We make no demographic or subgroup claim and run no equity audit. Equity is out of scope here by design.

\section*{Data and Code Availability}
The released English artifact is the build recipe and the fixed split indices only: the build script, the random seed, and the train, development, and test index files. We do not redistribute the underlying review text, which remains governed by the source CC BY-NC-SA license of the RateMyProfessor research release. Because the indices and the public source together reconstruct the corpus deterministically, any reconstruction is a derivative of CC BY-NC-SA content and inherits its terms: non-commercial research use, attribution to the source release, and redistribution under the same license. The benchmark harness, the gold-recovery script, and the bootstrap script are released alongside the recipe, available upon publication under the same license chain. The Spanish institutional corpus is not redistributed; it was accessed under Universidad de Guadalajara's research-use framework, and its availability is restricted under that framework.

\section*{Acknowledgments}
The Spanish protocol derives from the first author's prior thesis \citep{vega2026}. The institutional corpus was accessed under Universidad de Guadalajara's research-use framework.

\bibliography{references}

\appendix
\section{English corpus reproducibility}
\label{app:repro}
\textbf{Source.} RateMyProfessor research release \citep{he2020,he2022}, licensed CC BY-NC-SA. We use only the per-review star rating and comment text. We decode from the full release rather than the moral-foundations parquet, whose \texttt{student\_star} column is truncated to the 4.0 to 5.0 range and cannot yield a balanced sentiment label.

\textbf{Build.} Script \texttt{build\_rmp\_en\_sample.py}; output \texttt{rmp\_en\_sentiment\_\allowbreak{}sample.\{parquet,jsonl\}}; report \texttt{rmp\_en\_sample\_report.txt}.

\textbf{Pipeline.} We decode escaped bytes with \texttt{fix\_text}, then derive sentiment from \texttt{student\_star}: a review is positive if its rating is at least 4, negative if at most 2, and neutral otherwise. We drop star-null rows, comments under 15 characters, and exact duplicates. We then draw a uniform per-class reservoir sample over the full 9.54M-review file at seed 20260616, taking 15{,}000 per class, and make a stratified split.

\textbf{Result.} 45{,}000 comments, 15{,}000 per class; train 31{,}500, development 6{,}750, and test 6{,}750; zero exact duplicates and zero residual escapes; mean length 231 characters. The sentiment label is a deterministic function of the star rating, so it is consistent by construction. This is not a label-quality check; the human-label validation against EduRABSA appears in Section~\ref{sec:starlabel}. The 1{,}500-comment subset used for the cost-capped English LLM arms is the first 1{,}500 items of this same stratified 6{,}750-comment test split, so those LLM arms and the encoder arm are scored on identical gold labels for those items.

\section{Experiment status and supplementary analyses}
\label{app:b}
\subsection{Experiment status}
Completed analyses appear in Section~\ref{sec:expA} and Section~\ref{sec:expB}. Raw output is in \texttt{benchmark\_results.json}, the harness is \texttt{run\_paper2\_benchmark.py}, and all results use the single split with seed 20260616. The held-out gold labels are stored in the results file under \texttt{\_gold}, aligned one-to-one with the per-arm predictions, so every F1 value and every McNemar count is reproducible from the released artifact; the recovery and check script is \texttt{recover\_gold\_and\_bootstrap.py}. Completed: the Spanish arms TF-IDF, BETO, e5, Haiku, Opus, and the open-weights Qwen2.5-1.5B, each zero- and few-shot; the English arms on the 45{,}000-corpus splits; McNemar counts; and the EduRABSA star-versus-human agreement at Cohen's kappa 0.66, 82\% accuracy, $n{=}3000$.

\subsection{Descriptive bootstrap confidence intervals}
\label{app:boot}
We summarize the headline weighted-F1 differences with 95\% percentile bootstrap intervals over 2{,}000 resamples of the held-out items, computed on the stored predictions and recovered gold labels (Table~\ref{tab:boot}). These are descriptive: an interval that includes zero indicates no descriptive separation, and no significance is claimed. Where the English arms are scored on different sample sizes their per-arm intervals overlap: the e5 encoder reaches weighted F1 0.729 with interval $[0.718, 0.740]$ on the full 6{,}750-comment split, while Haiku few-shot reaches 0.733 with interval $[0.710, 0.756]$ on the 1{,}500-comment subset, so the encoder lead over the cost-capped LLM arms is indicative rather than separable.

\begin{table}[t]\centering\footnotesize\setlength{\tabcolsep}{2.5pt}
\begin{tabular}{l r l}
\toprule
Comparison & $\Delta$wF1 & 95\% CI \\
\midrule
ES sent: Haiku-0 $-$ Opus-0 & +0.040 & $[0.007,\ 0.075]$ \\
ES sent: TF-IDF $-$ BETO & -0.024 & $[-0.078,\ 0.032]$ \\
ES sent: BETO $-$ e5 & -0.021 & $[-0.069,\ 0.027]$ \\
ES them: BETO $-$ Opus-few & -0.142 & $[-0.218,\ -0.079]$ \\
ES them: BETO $-$ e5 & +0.004 & $[-0.055,\ 0.065]$ \\
EN sent: e5 $-$ TF-IDF & +0.054 & $[0.043,\ 0.066]$ \\
EN sent: Haiku-f $-$ Opus-f & +0.010 & $[-0.006,\ 0.026]$ \\
\bottomrule
\end{tabular}
\caption{\textbf{Descriptive bootstrap confidence intervals} for headline weighted-F1 differences (95\% percentile, 2{,}000 resamples). An interval covering zero indicates no descriptive separation; no significance is claimed. ES arms use the 233-comment split; EN LLM arms use the 1{,}500-comment subset.}
\label{tab:boot}
\end{table}

\subsection{LIME auditability check}
\label{app:lime}
As an auditability sanity check we apply LIME \citep{ribeiro2016} to held-out Spanish predictions from the classical TF-IDF arm and the e5 encoder arm. The output (\texttt{lime\_explanations.txt}, produced by \texttt{lime\_audit.py}) shows feature words and labels only, with personal names masked, so no comment is displayed in full. Local explanations align with the annotation guide. On the thematic task, \textit{grosero} and \textit{amable} drive the \textit{interaction} class, \textit{evaluaci\'on} and \textit{calificaci\'on} drive \textit{evaluation}, and \textit{faltaba} and \textit{tarde} drive \textit{attendance/engagement}. On sentiment, \textit{excelente} and \textit{muy} drive the positive class, while \textit{no}, \textit{falta}, and \textit{grosero} drive the negative class. The neutral class leans on hedging cues such as \textit{pero} and \textit{sin embargo}, consistent with its lower per-class F1. These attributions are inspectable per prediction, the auditability that the classical arms provide and the prompt-bound LLM arms do not.
\typeout{Label(s) may have changed. Rerun to get cross-references right.}
\end{document}